\pdfoutput=1

\documentclass[11pt]{article}

\usepackage{acl}

\usepackage{times}
\usepackage{latexsym}

\usepackage[T1]{fontenc}

\usepackage[utf8]{inputenc}

\usepackage{microtype}
\usepackage{graphicx}                                                           
\usepackage{float}
\usepackage{inconsolata}
\usepackage{booktabs}
\usepackage{amsmath} 
\usepackage{makecell}
\usepackage{colortbl}  
\usepackage{xcolor}
\usepackage{hhline}
\usepackage{enumerate}
\usepackage{booktabs}
\usepackage{tabularx}
\usepackage{tcolorbox}
\tcbuselibrary{breakable}
\usepackage{listings}
\definecolor{highlightyellow}{RGB}{255,255,153}
\usepackage{algorithm}
\usepackage{algpseudocode}

\usepackage{graphicx}
\usepackage{bbding}
\title{EPI-SQL: Enhancing Text-to-SQL Translation with Error-Prevention Instructions}

\author{Xiping Liu ,  Zhao Tan \\
        Jiangxi University of Finance and Economics}

\begin{document}
\maketitle
\begin{abstract}

The conversion of natural language queries into SQL queries, known as Text-to-SQL, is a critical yet challenging task. This paper introduces EPI-SQL, a novel methodological framework leveraging Large Language Models (LLMs) to enhance the performance of Text-to-SQL tasks. EPI-SQL operates through a four-step process. Initially, the method involves gathering instances from the Spider dataset on which LLMs are prone to failure. These instances are then utilized to generate general error-prevention instructions (EPIs). Subsequently, LLMs craft contextualized EPIs tailored to the specific context of the current task. Finally, these context-specific EPIs are incorporated into the prompt used for SQL generation. 
EPI-SQL is distinguished in that it provides task-specific guidance, enabling the model to circumvent potential errors for the task at hand. Notably, the methodology rivals the performance of advanced few-shot methods despite being a zero-shot approach. An empirical assessment using the Spider benchmark reveals that EPI-SQL achieves an execution accuracy of 85.1\%, underscoring its effectiveness in generating accurate SQL queries through LLMs. The findings indicate a promising direction for future research, i.e. enhancing instructions with task-specific and contextualized rules, for boosting LLMs' performance in NLP tasks.

\end{abstract}

\section{Introduction}
Text-to-SQL is a task in natural language processing (NLP) that aims to automatically generate structured query language (SQL) queries from natural language text. This task enables users to access databases without requiring SQL knowledge or familiarity with the database schema, thus facilitating the work of data analysts and software developers who need to write complex SQL queries. Text-to-SQL has attracted significant interest from both industry and academia in recent years \cite{wang2020rat,choi2021ryansql,zhao2022importance}.

With the rapid progress of Large Language Models (LLMs), the research areas of NLP are being revolutionized \cite{zhao_SurveyLargeLanguage_2023}. LLMs can now serve as a general-purpose language task solver (to some extent), and they have shown impressive performance in a series of NLP tasks, e.g., arithmetics, symbolic reasoning \cite{kojima2022large}, disambiguation QA, movie recommendation, etc. \cite{suzgun_ChallengingBIGBenchTasks_2022}. 

Numerous studies have explored the application of LLMs for Text-to-SQL, and a number of LLM-based methods have been proposed \cite{rajkumar2022evaluating, liu2023comprehensive, pourreza2023din, gao2023text}. These methods fall into two categories: zero-shot and few-shot. In zero-shot settings, LLMs are tasked with generating SQL queries based solely on task-descriptive instructions. Conversely, few-shot prompting involves providing LLMs with a small number of demonstrations, facilitating task completion through In-Context Learning (ICL) \cite{brown2020language, radford2019language}. The few-shot approach has attracted considerable attention due to its ability to provide models with additional context through the use of examples \cite{chen2023teaching, pourreza2023din, ni2023lever, liu2022makes, nan2023enhancing, guo2023case, gao2023text}. 
In contrast, the zero-shot approach has not received as much research interest. We believe that the zero-shot approach remains underexplored and holds significant untapped potential. Therefore, the focus of this paper is to enrich zero-shot prompts with specially designed instructions, aiming to unlock the full potential of this approach. 

\begin{figure*}[!ht]
    \centering
    \includegraphics[width=\linewidth,scale=1]{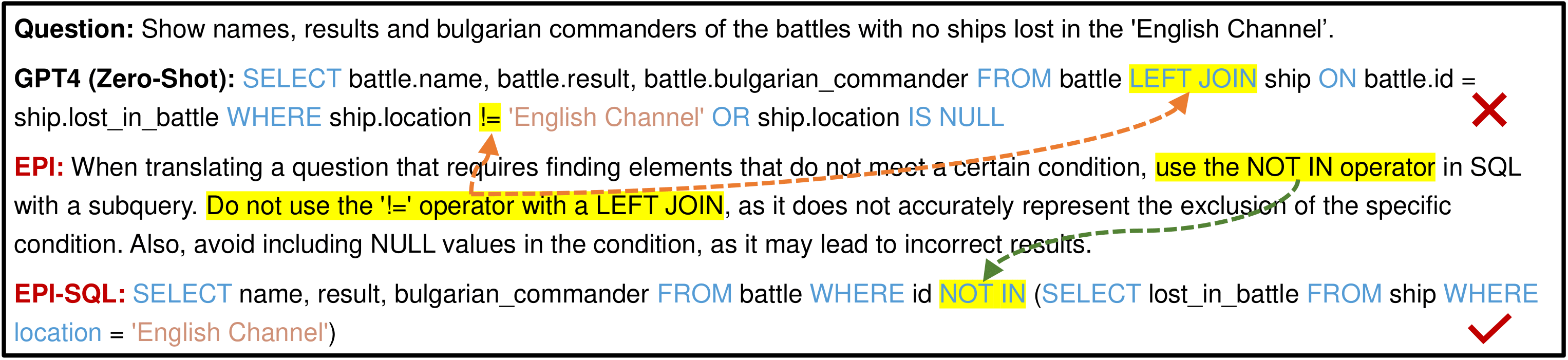}
    \caption{An example of EPI and answers generated by EPI-SQL. The orange line represents the connection of EPI and potential errors, and the green line indicates the connection of correct answer and EPI.}
    \label{fig: examples_for_EPI-SQL}
\end{figure*}

Typically, an instruction within the context of a Text-to-SQL task is simply a sentence that outlines the task, such as "Please translate this question into a SQL query." However, such instructions are suboptimal for two primary reasons: 
(1) Limited information. Traditional instructions merely activate the model's Text-to-SQL capabilities without providing additional useful information, thereby only scratching the surface of LLMs' capabilities; 
(2) Lack of context awareness. The same instruction is applied to all questions without adaptation to the varying context of different questions. It is more desirable to provide instructions that are tailored to individual questions.

To address these issues, this paper proposes to enhance zero-shot prompts with \emph{Error-Prevention Instructions} (EPIs). As shown in Figure \ref{fig: examples_for_EPI-SQL}, the EPI is designed to furnish precise and valuable information for each Text-to-SQL task. In comparison to traditional instructions, the EPI provides comprehensive information that includes accurate guidance for the current task while also prompting LLMs to avoid potential errors.

With EPI, we introduce EPI-SQL, a novel zero-shot method for Text-to-SQL. The basic idea behind EPI-SQL is to derive insights from past mistakes. Through an analysis of Text-to-SQL outcomes on LLMs, we observed that a LLM may make similar mistakes when encountering comparable questions or fail under analogous circumstances. To effectively leverage these errors, we initially compiled a set of error-prone instances in the Text-to-SQL task. Then, a number of prevention rules, i.e., EPIs, are drawn from these instances. When presented with a question, the method finds the EPIs most relevant to the current question, synthesizes the EPIs and proposes a contextualized EPI that cater to the context of the current task. This method enables the LLM to anticipate and avoid potential errors, ultimately leading to improved results.

Through experimentation, we have demonstrated the effectiveness of the EPI-SQL. On the Spider dataset \cite{yu2018spider}, EPI-SQL achieved an execution accuracy of \textbf{85.1\%} and a test suite accuracy of \textbf{77.9\%}, outperforming the existing state-of-the-art systems. It is worth noting that these results were obtained in a zero-shot scenario, underscoring the substantial potential of instruction-enhancing techniques, which has often been overlooked in previous research.

In summary, we make the following contributions in this work:
\begin{itemize}
     \item We present an enhancement to zero-shot prompts through the incorporation of \emph{Error-Prevention Instructions} (EPIs). The EPI provides comprehensive information that delivers precise guidance for the current task while simultaneously prompting large language models to circumvent potential errors.
    
   \item We present EPI-SQL, an innovative approach to Text-to-SQL tasks that incorporates EPIs. This method enables LLMs to learn not only from past errors but also to proactively anticipate potential future errors, ultimately enhancing overall performance results.
    
    \item We conduct a comprehensive set of experiments and find that EPIs are highly beneficial for the Text-to-SQL task. In fact, our method achieves high performance in a zero-shot setting, demonstrating the significant potential of EPIs and instruction-enhancing techniques.
\end{itemize}

\section{Methodology}
\subsection{Problem Description}
\begin{figure*}[!ht]
    \centering
     \includegraphics[width=\linewidth,scale=1]{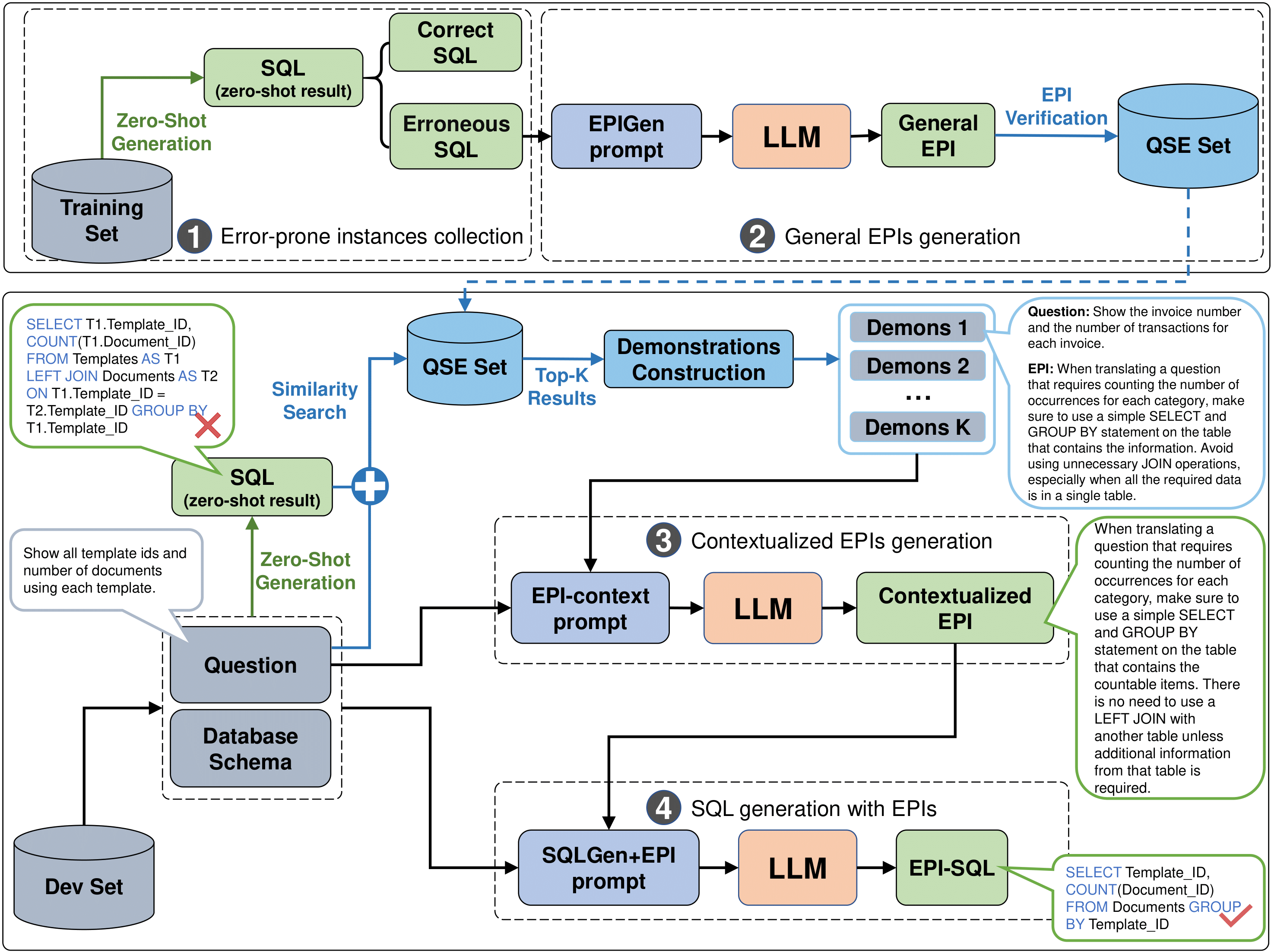}
    \caption{The framework of our method.}
    \label{fig:Framework}
\end{figure*}

Text-to-SQL is a task that maps a natural language question $Q$ to a SQL query given a database schema. In this work, we investigate the use of LLMs for Text-to-SQL.
Prompts in LLMs-based applications act as the essential input that steers the model towards producing contextually relevant and accurate outputs. A basic strategy of prompting an LLM is zero-shot prompting, where the model is provided with a task description, also known as instruction, $I$, and expected to generate the desired output. An LLM $\mathcal{M}$ estimates a probability distribution over SQL queries $y$, allowing us to generate queries token by token. Thus, the zero-shot Text-to-SQL generation with LLMs can be formulated as follows:
\begin{equation}
\resizebox{1\linewidth}{!}{$
P_{\mathcal{M}}(y \mid x)=\prod_{i=1}^{|y|} P_{\mathcal{M}}\left(y_i \mid \operatorname{prompt} \left(Q, D, I \right), y_{<i}\right),
$}
\end{equation}
where $x$ is the current task's input, $y$ is the output; the question $Q$, database schema $D$ and instruction $I$ make up a zero-shot prompt. $y_{<i}$ is the prefix of the SQL query $y_i$ and $P_{\mathcal{M}}\left(y_i \mid \cdot\right)$ is the conditional probability of the $i$-th token in the SQL query $y$ given the prefix $y_{<i}$ and the prompt.

\subsection{Overview of Our Method}
In prompting-based methods,  instructions play a vital role in obtaining optimal results from LLMs. 
Current  approaches often rely on uniform, static instructions that do not account for the nuances across different contexts, leading to suboptimal results, particularly when dealing with complex or contextually rich queries.
In our research, we introduce a novel strategy that incorporates \emph{dynamic}, \emph{contextualized} instructions, called \emph{error-prevention instructions} (EPIs). These instructions are adaptive, providing tailored guidance that aligns with the context of the current task. Moreover, these instructions are designed to preemptively address potential errors. 

As shown in Figure \ref{fig:Framework}, our method comprises four main components: error-prone instances collection, general EPIs generation, contextualized EPIs generation, and SQL generation with EPIs. We will discuss these components in detail in the following sections.

\begin{figure*}[!t]
    \centering
     \includegraphics[width=\linewidth,scale=1]{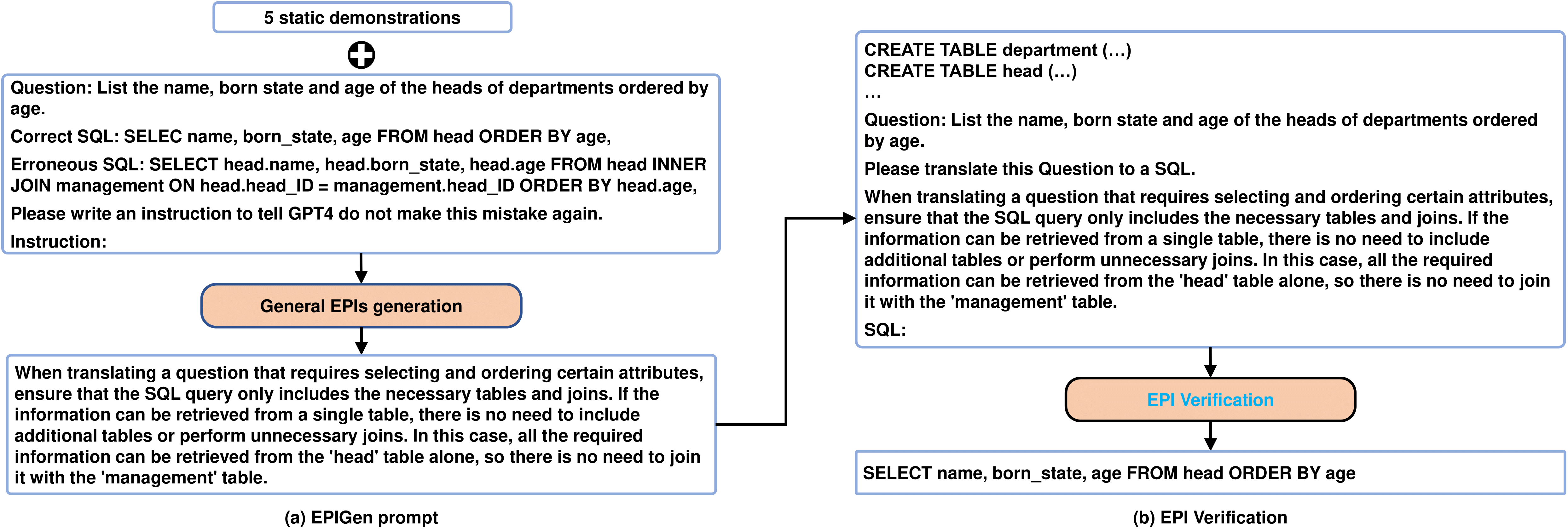}
    \caption{Examples of the input and output of prompts used to construct the QSESet.}
    \label{fig:prompts in training set}
\end{figure*}

\begin{figure*}[!t]
    \centering
     \includegraphics[width=\linewidth,scale=1]{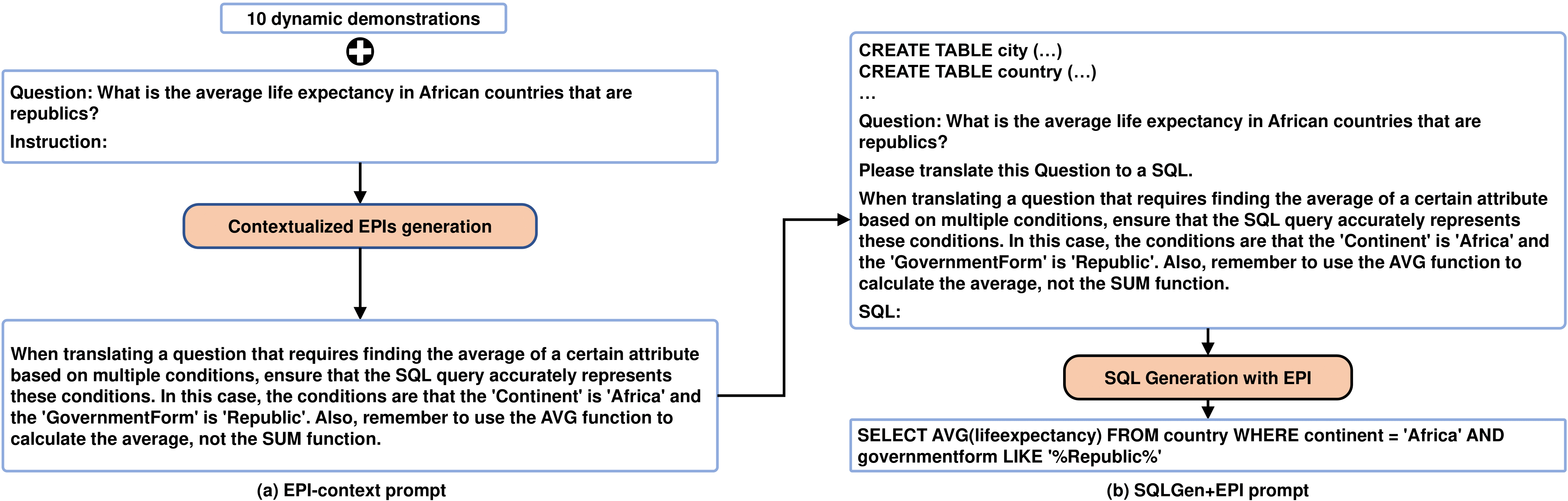}
    \caption{Examples of the input and output of prompts used to generate contextualized EPI and EPI-SQL.}
    \label{fig:prompts in dev set}
\end{figure*}

\subsection{Error-prone Instances Collection}
In our study, we employed a zero-shot prompting technique on the training set of the Spider dataset \cite{yu2018spider} to produce a response for each instance. Subsequently, the generated responses were compared with the gold-standard answers, and the examples that were not generated accurately were collected. These examples highlight the situations where the LLMs might make mistakes.

\subsection{General EPIs Generation}
After the error-prone instances are collected, they are fed into LLMs. The LLMs are tasked with formulating a set of instructions aimed at circumventing these errors. These instructions, referred to as error-prevention instructions (EPIs), are generated by an LLM using a prompt, called \textbf{EPIGen prompt}, as shown in Figure \ref{fig:prompts in training set} (a).

To further refine the EPIs, we employ a process of \emph{EPI-verification}, as depicted in Figure \ref{fig:prompts in training set} (b). For each instance $S$ where an error might occur, an EPI $E_S$ is generated using the EPIGen prompt. We then apply the EPI $E_S$ to the problematic instance $S$ to test its effectiveness. If the EPI  $E_S$  successfully guides the LLM to  handle the instance $S$ correctly, $E_S$ along with the question $Q_S$ in $S$, and the erroneous SQL $SQL_S$ generated for $S$ will be recorded in a set $QSESet$ (Question-SQL-EPI set). 

\subsection{Contextualized EPI Generation}
The previously generated EPIs are broad, encompassing a wide range of potential errors, and may not be directly applicable to the task at hand. Thus, our objective at this step is to derive EPIs that are contextualized—that is, EPIs specifically pertinent to the task currently being addressed. This is achieved by engaging the LLM with a specific prompt called the \textbf{EPI-context prompt}, as depicted in Figure \ref{fig:prompts in dev set} (a). 

The EPI-context prompt includes a set of carefully selected demonstrations, which direct the LLM to associate queries with suitable EPIs, thus enabling the generation of task-specific EPIs. 
The development of these demonstrations proceeds as follows: For a given Text-to-SQL task with a question $Q$, our method initially produces an answer $SQL_Q$ using a straightforward zero-shot prompt. Subsequently, we identify the top-$k$ instances most akin to the current task by searching within the $QSESet$. Note that, when computing the similarities, both the question $Q$ and the answer $SQL_Q$ are taken into account. Finally, a set of demonstrations is assembled, with each one drawn from an instance, consisting of the instance's question paired with a corresponding EPI.

Using the EPI-context prompt, which encompasses  the context of the current task and the set of selected demonstrations, the LLM generates an EPI that is tailored to the task at hand. 

\subsection{SQL Generation using EPI}
In this phase, EPIs are harnessed to facilitate the generation of SQL. To this end, we introduce what we term the \textbf{SQLGen+EPI prompt}, as illustrated in Figure \ref{fig:prompts in dev set} (b). The SQLGen+EPI prompt comprises three constituents: the question, the schema of the database, and the instructions, which notably include the EPI.

Note that we do not use demonstrations in this prompt. In other words, it is a zero-shot prompt other than a few-shot prompt. We will demonstrate that this zero-shot prompting method achieves remarkable performance not inferior to other methods with demonstrations.

\section{Experiments}
In this section, we carry out a series of experimental studies to investigate the effectiveness of the methods proposed in this work.

\subsection{Experiment Settings}
\textbf{Models.} In the experiments, we used a powerful and publicly accessible LLM: GPT-4. We perform greedy decoding at temperature $\tau$ = 0 to ensure stable output. In similarity computation, we use text-embedding-ada-002 model to get the embedding of question and SQL.

~\\\textbf{Datasets.} We conducted experiments on the standard Spider dataset\cite{yu2018spider}, a large-scale cross-domain Text-to-SQL benchmark containing 7000 training samples across 146 databases and 1034 evaluation samples across 20 databases.

~\\\textbf{Evaluation Metrics.} We use Execution Accuracy (EX) as the evaluation metric for all experiments, which measures the percentage of system predictions leading to the gold execution result. We also adopt test-suite accuracy (TS) \cite{zhong2020semantic} as an evaluation metric. TS could achieve high code coverage from a distilled test suite of the database, and it is also based on execution results.

~\\\textbf{Baselines.}
We compare our methods with the following baselines:

\noindent (1) \emph{Zero-shot + ChatGPT} \cite{liu2023comprehensive}, which works on ChatGPT under a zero-shot setting.

\noindent (2) \emph{Coder-Reviewer} \cite{zhang2022coder}, which generates and selects SQL queries based on their likelihood.

\noindent (3) \emph{MBR-Exec} \cite{shi2022natural}, which generates and selects SQLs with the most common execution result.

\noindent (4) \emph{PICARD} \cite{scholak2021picard}, which constrains auto-regressive coders of language models through incremental parsing.

\noindent (5) \emph{RASAT } \cite{qi2022rasat}, which introduces relation-aware self-attention into transformer models and utilizes constrained auto-regressive decoders.

\noindent (6) \emph{C3-Prompt} \cite{dong2023c3} a systematic zero-shot GPT-based Text-to-SQL approach.

\noindent (7) \emph{LEVER} \cite{ni2023lever}, which generates SQL queries with Codex and selects the one with the highest scores.

\noindent (8) \emph{RESDSQL} \cite{li2023resdsql}, a ranking-enhanced encoding and skeleton-aware decoding framework.

\noindent (9) \emph{Self-Debug} \cite{chen2023teaching}, which prompts LLMs to debug SQL with explanations.

\noindent (10) \emph{PDS-Prompt} \cite{nan2023enhancing}, which selects few-shot demonstrations in terms of diversity and similarity.

\noindent (11) \emph{DIN-SQL} \cite{pourreza2023din}, which decomposes text-to-SQL tasks into four subtasks and prompts LLMs to solve them sequentially.

~\\\textbf{Other Details.}
When finding the erroneous instances, we used the GPT-4 model with zero-shot prompts on the training set of Spider dataset. This approach yielded an accuracy of 84.5\%, with a total of 1,083 out of 7000 instances being inaccurately classified. Subsequently, 529 instances were left after verification, and used to construct the QSESet.

During demonstration selection, we use the text-embedding-ada-002 model to obtain the embeddings of the questions and SQLs, and then select the top-$10$ demonstrations with the highest cosine similarities.

\subsection{Main Result}
Table~\ref{tab:main experiment} shows the performance of our method and other baselines on the Spider development set. It can be seen that our method with GPT-4 model achieves the best performance on this benchmark. 

Among the various methods evaluated, C3-prompt \cite{dong2023c3} employs a zero-shot pipeline, DIN-SQL \cite{pourreza2023din} adopts a few-shot pipeline, DPS-prompt \cite{nan2023enhancing} is a single-round few-shot method, and Self-Dedug \cite{chen2023teaching} uses a multi-round few-shot strategy.
Our EPI-SQL does not use demonstrations for generating SQL queries. In this regard, it aligns with a zero-shot approach. Nevertheless, it uniquely employs instances as a foundation for generating instructions, thereby differentiating it from conventional zero-shot methodologies. This distinctive trait of EPI-SQL suggests potential implications beyond traditional zero-shot and few-shot prompting methods. 

\begin{table}[ht]
\centering
\resizebox{\linewidth}{!}{
\begin{tabular}{l c c}
\toprule[1pt]
\textbf{Methods} & \textbf{EX} & \textbf{TS} \\
\midrule[0.5pt]
Zero-shot + ChatGPT \cite{liu2023comprehensive}   & 70.1 & 60.1 \\
Coder-Reviewer + CodeX\cite{zhang2022coder}             &  74.5 & -  \\
MBR-Exec \cite{shi2021learning}                     &  75.2 & -  \\
T5-3B + PICARD \cite{scholak2021picard}          & 79.3 & 69.4\\
RASAT + PICARD \cite{li2023graphix}            & 80.5 & 70.3 \\
C3-prompt + ChatGPT \cite{li2023graphix}            & 81.8 & - \\
LEVER + CodeX \cite{ni2023lever}                         &  81.9 & -  \\
RESDSQL-3B + NatSQL \cite{li2023resdsql}                   & 84.1 & 73.5 \\ 
Self-Debug + CodeX \cite{chen2023teaching}       &  84.1  & - \\
PDS-pormpt + CodeX \cite{nan2023enhancing}            &    84.4 & - \\
DIN-SQL + GPT-4 \cite{pourreza2023din}          &  \textbf{84.8}  &  \textbf{74.2}  \\
\midrule[0.5pt]
EPI-SQL + GPT-4 (Ours)  & \textbf{85.1} & \textbf{77.9} \\
\bottomrule[1pt]
\end{tabular}
}
\caption{Execution accuracy (EX) and test-suite accuracy (TS) on the Spider development set.}
\label{tab:main experiment}
\end{table}

\subsection{Ablation Study}
In this section, we present the results of the ablation study, as shown in  Table \ref{tab:Ablation Study}. According to the difficulty of generating SQL, the Spider dataset can be divided into four subsets: easy, medium, hard, and extra-hard \cite{yu2018spider}.  EPI could significantly improve the performance of Text-to-SQL tasks across levels of complexity from medium to extra-hard. While easy-level tasks typically encompass single-table queries, errors in these tasks often arise from a deficiency in external domain knowledge rather than the SQL structure itself. This unexpected challenge can lead to bottlenecks for LLMs when addressing easy-level tasks.

Our ablation study assessed the efficacy of our method both with and without its critical elements, such as EPI-verification, different types of similarities, and EPI. SQL-similarity and Question-similarity refer to the similarities between SQLs and questions, respectively, which are used in demonstration selection. It should be noted that without EPI, our method is reduced to a basic zero-shot prompting approach.

As shown in Table \ref{tab:Ablation Study}, excluding any component of our method can cause a decrease in the overall performance.
The absence of an EPI-verification mechanism has the most profound effect because, during the construction of the QSESet, the inability to weed out invalid EPIs leads to contamination of the set, which in turn negatively influences the generation of EPIs for the current task. The similarity between questions is of greater importance since questions often include explicit information pertinent to SQL, which is essential for the successful completion of Text-to-SQL tasks.

\begin{table}[h]
\centering
\resizebox{\linewidth}{!}{
\begin{tabular}[!]{lccccc}
\toprule[1pt]
\multicolumn{6}{c}{\textbf{Execution accuracy}} \\
\midrule[0.5pt]
\textbf{Technique} & \textbf{Easy} & \textbf{Medium} & \textbf{Hard} & \textbf{Extra-hard} & \textbf{EX} \\
\midrule[0.5pt]
EPI-SQL  &   \textbf{93.1}   &   \textbf{90.5}    &   \textbf{81.8} &  \textbf{62.0}  & \textbf{85.1}  \\
w/o EPI-verification      & 92.3 &  86.7 &  77.3  & 57.8  &  81.8   \\
w/o SQL-similarity       & \textbf{93.1} &  87.2 &  78.4  & 60.8  &  82.9   \\
w/o Question-similarity  & 91.5 &  88.1 &  77.8  & 58.4  &  82.4   \\
w/o EPI (simple zero-shot)   & \textbf{93.1} &  86.5 &  78.4  & 59.0   &  82.3   \\
\midrule[1pt]
\midrule[1pt]

\multicolumn{6}{c}{\textbf{Test-suit accuracy}}  \\
\midrule[0.5pt]
\textbf{Technique} & \textbf{Easy} & \textbf{Medium} & \textbf{Hard} & \textbf{Extra-hard} & \textbf{EX} \\
\midrule[0.5pt]
EPI-SQL  &   \textbf{91.9}   &   \textbf{86.3}    &   \textbf{68.6} &  \textbf{44.0}  & \textbf{77.9}  \\
w/o EPI-verification      & 90.7 &  78.4 &  64.2  & 33.1  &  71.7   \\
w/o SQL-similarity       & \textbf{91.9} &  83.1 &  63.6  & 42.8  &  75.4   \\
w/o Question-similarity  & 90.7 &  84.0 &  64.8  & 42.2  &  75.6   \\
w/o EPI (simple zero-shot)   & 91.5 &  78.1 &  65.3  & 34.9   &  72.2   \\
\bottomrule[1pt]
\end{tabular}}
\caption{Performance of our method with and without each step.}
\label{tab:Ablation Study}
\end{table}

\section{Text-to-SQL Biases Analysis}

\begin{figure*}[!ht]
    \centering
    \includegraphics[width=\linewidth,scale=1]{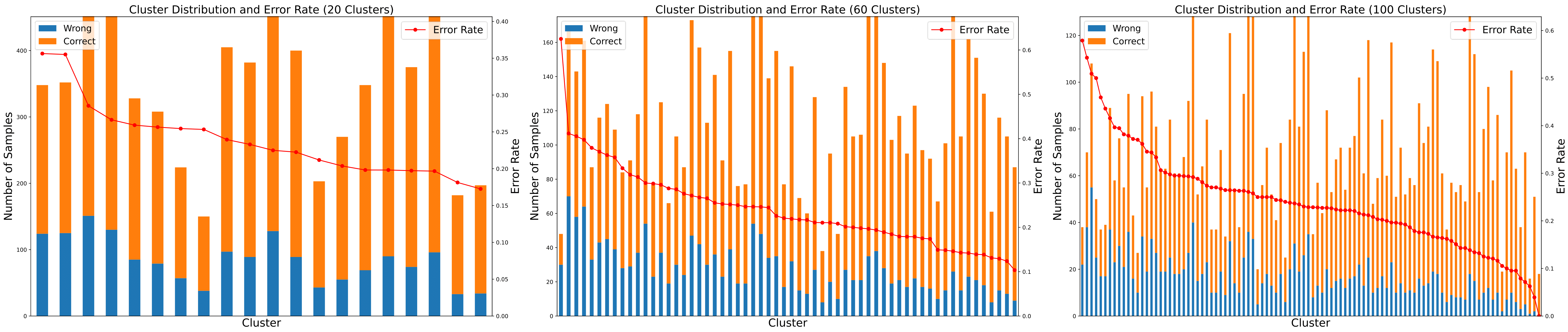}
    \caption{Error distribution and error rate for each question cluster, where cluster = 20,60,100.}
    \label{fig:bias distribution of question}
\end{figure*}

\begin{figure*}[!ht]
    \centering
    \includegraphics[width=\linewidth,scale=1]{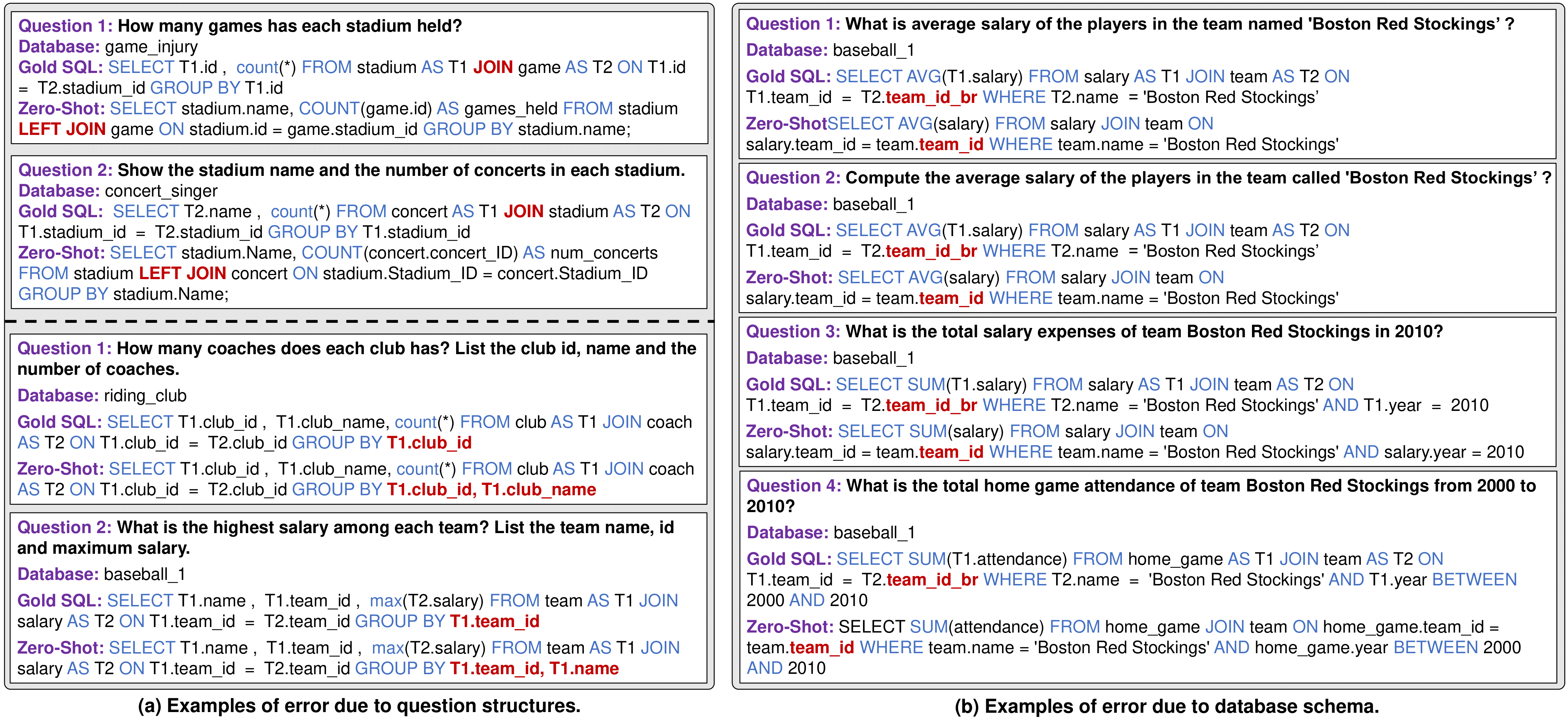}
    \caption{Examples of error.}
    \label{fig:Examples of bias due to question and schema}
\end{figure*}

LLMs exhibit a tendency for systematic mistakes in specific tasks, which can be considered as \emph{biases} \cite{dong2023c3}. The biases are usually due to the uneven distribution of data the model is exposed to during training or to some inherent characteristic of the model architecture. 
Biases exist in Text-to-SQL tasks. An example of the biases is that LLMs tend to repeat the same mistakes across analogous Text-to-SQL tasks. For instance, the model might frequently use 'LEFT JOIN' instead of a simple 'JOIN'. Such biases can adversely affect the performance of Text-to-SQL.

In this section, we aim to identify patterns and factors contributing to the model's biased behavior on Text-to-SQL. This analysis will enhance our understanding of the model's limitations and inform the development of more robust and equitable Text-to-SQL systems.

We first analyzed the questions. The 7,000 questions from the Spider \cite{yu2018spider} training set were clustered using the $k$-means algorithm. 
Figure \ref{fig:bias distribution of question} illustrates the distribution of correct and incorrect samples (zero-shot results) and the error rates when the number of clusters ($k$) is set to 20, 60, and 100. We can see that the LLMs indeed exhibit higher error rates for specific clusters of questions, with this pattern becoming increasingly pronounced as the cluster count rises. This indicates that the intrinsic features of the questions might be a significant determinant of the biases perceived in the model's performance.
Figure \ref{fig:Examples of bias due to question and schema} (a) shows a series of Text-to-SQL tasks with similar errors attributed to their analogous structure within the questions.

\begin{figure}[!t]
    \centering
    \includegraphics[width=\linewidth,scale=1]{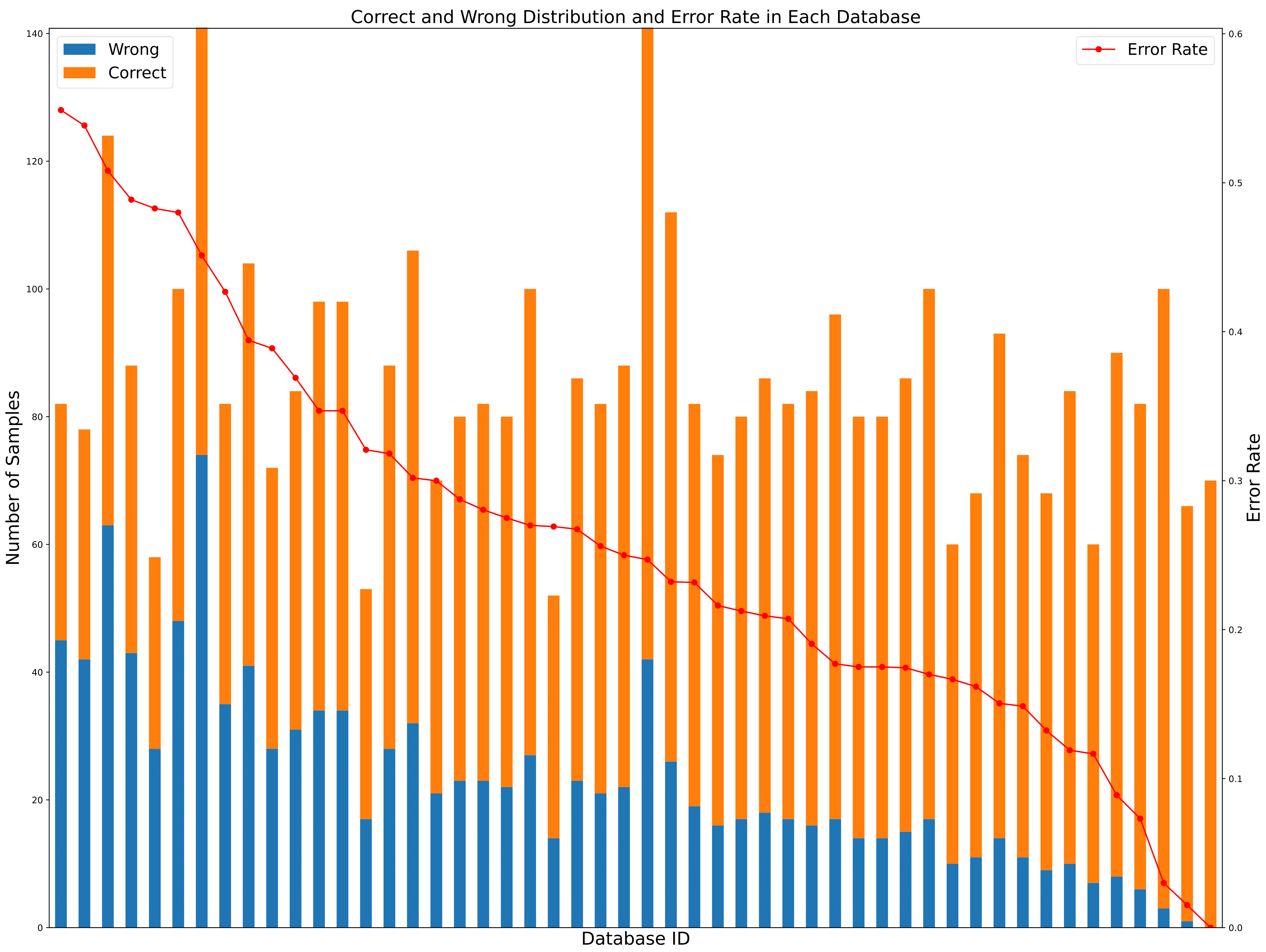}
    \caption{Error distribution and error rates across databases.}
    \label{fig:bias of schema}
\end{figure}

Although there is also a difference in the error distribution across different databases, as depicted in Figure \ref{fig:bias of schema}, this is primarily attributed to the database schema itself, such as atypical column names or the complex interrelations among tables. 

For example, as shown in Figure \ref{fig:Examples of bias due to question and schema} (b), the model generated table join errors on these tasks because they did not recognize the foreign key in the \textit{Team} table, "team\_id\_br", an unconventional column name. Specifically, the \textit{baseball\_1} database is quite complex, with 26 tables and 341 columns, in which some column names are ambiguous, leading to low accuracy. We think that these errors stem not from the inherent bias of the model but rather from the difficulty in schema linking.


\section{Related Work}
Text-to-SQL aims to simplify the process of accessing data in relational databases for non-expert users. Researchers have made impressive achievements in this task by designing models \cite{wang2020rat,cai2021sadga,li2023graphix,qi2022rasat, li2023resdsql} or fine-tuning pre-trained models \cite{yugrappa,shi2021learning,scholak2021picard}. 

LLMs have demonstrated impressive code generation abilities without fine-tuning \cite{chen2021evaluating, chowdhery2022palm, zhao2022importance, athiwaratkun2022multi}. 
A series of research studies have been done to investigate the capacity of LLMs on Text-to-SQL.
\cite{rajkumar2022evaluating, liu2023comprehensive} studied the efficacy of Text-to-SQL on various LLMs. They explored the impact of prompt structure, number of few-shot demonstrations, and other factors on the outcomes using zero-shot and few-shot prompting. 

The rapid development of prompting-based methods has led to the proposal of numerous effective prompting principles. For example, CoT prompting \cite{kojima2022large} is proposed to improve LLMs' reasoning ability by producing intermediate steps before predicting a final answer; Self-Consistency \cite{wang2022self} mitigates the phenomenon of randomness in the output of LLMs by voting on the diversity of results and selecting the best one. For Text-to-SQL, these prompting enhancement methods are equally effective. Self-Debug\cite{chen2023teaching} employing CoT prompting to obtain the question explanation and generates the initial SQL, then instruct LLMs to debug the SQL. Coder-Reviewer \cite{zhang2022coder}, MBE-Exec \cite{shi2022natural} and LEVER \cite{ni2023lever} utilizing consistency principles to choose the optimal one from multiple candidate results. MBE-Exec \cite{shi2022natural} selects the SQL with the most common execution result, Coder-Reviewer \cite{zhang2022coder} selects the SQL considering both the likelihood of the predicted SQL; LEVER \cite{ni2023lever} selects the SQL with the highest score, which represents the probability that the SQL is correct and is calculated based on the question, the SQL and the execution results. 

LLMs-based Text-to-SQL methods include pipelined, single-round, and multi-round methods.
Pipeline methods improve performance by decomposing a Text-to-SQL task to reduce its complexity. DIN-SQL \cite{pourreza2023din} breaks down the task into four sub-tasks and applies few-shot learning, while C3-prompt \cite{dong2023c3} divides it into two parts using zero-shot learning. Single-turn approaches focus on input representation and demonstration selection, with DPS-prompt \cite{nan2023enhancing} using a similarity-diversity sampling strategy for demonstration selection. Multi-round methods involve interaction with external information. Self-debug \cite{chen2023teaching}, for example, continually makes new demands on the output of the previous round to improve performance. However, these methods have not considered the potential of instructions in Text-to-SQL.

\section{Conclusion}
In this study, we introduce EPI-SQL, a method that employs zero-shot prompting with Large Language Models (LLMs) for Text-to-SQL tasks, notable for its delivery of performance comparable to leading few-shot techniques. 
EPI-SQL is novel in that it actively identifies known errors and distills error-prevention instructions to refine the prompts used. Its notable feature lies in its ability to generate practical, task-specific guidance that enables the model to circumvent potential errors relevant to the current task. In the Spider benchmark, EPI-SQL achieves an impressive 85.1\% execution accuracy, showcasing the great potential to prompt LLMs with effective instruction.

\section*{Limitations}
This work has certain limitations. Firstly, the validity of EPIs may not be consistent across different LLMs due to varying error patterns. Additionally, the database schema terms may impede the similarity-based demonstration selection strategy, which could be investigated in future work.
\bibliography{custom}
\bibliographystyle{acl_natbib}

\appendix
\onecolumn
\label{sec:appendix}

\section{Prompts}
In this section, we show the prompts used in this paper.

\subsection{Zero-shot Prompt}
Zero-shot prompt is used to generate zero-shot results.

\begin{lstlisting}[xleftmargin=0pt]
CREATE TABLE department (
Department_ID INT,
Name TEXT,
Creation TEXT,
Ranking INT,
Budget_in_Billions REAL,
Num_Employees REAL,
PRIMARY KEY (Department_ID)
);
CREATE TABLE head (
head_ID INT,
name TEXT,
born_state TEXT,
age REAL,
PRIMARY KEY (head_ID)
);
CREATE TABLE management (
department_ID INT,
head_ID INT,
temporary_acting TEXT,
PRIMARY KEY (department_ID,head_ID),
FOREIGN KEY(head_ID) REFERENCES head(head_ID),
FOREIGN KEY(department_ID) REFERENCES department(Department_ID)
);

Question: How many heads of the departments are older than 56 ?
Please translate question to SQL based on given dataset schema.
SQL:
\end{lstlisting}

\subsection{EPIGen Prompt}
The EPIGen prompt generates an EPI by contrasting the erroneous SQL with the correct one.

\begin{lstlisting}[xleftmargin=0pt]
Question: Find the title of course that is provided by both Statistics and Psychology departments.,
Correct SQL: SELECT title FROM course WHERE dept_name  =  'Statistics' INTERSECT SELECT title FROM course WHERE dept_name  =  'Psychology',
Erroneous SQL: SELECT title FROM course WHERE dept_name IN ('Statistics', 'Psychology') GROUP BY title HAVING COUNT(DISTINCT dept_name) = 2
Please write an instruction to tell GPT4 do not make this mistake again.
Instruction: When translating a question that requires finding common elements between two categories, use the INTERSECT operator in SQL. Do not use the IN operator with GROUP BY and HAVING COUNT, as it does not accurately represent the intersection of the two categories.

Question: Which studios have never worked with the director Walter Hill?,
Correct SQL: SELECT Studio FROM film EXCEPT SELECT Studio FROM film WHERE Director  =  "Walter Hill",
Erroneous SQL: SELECT Studio FROM film WHERE Director != 'Walter Hill' GROUP BY Studio
Please write an instruction to tell GPT4 do not make this mistake again.
Instruction: When translating a question that requires finding elements that have never been associated with a certain condition, use the EXCEPT operator in SQL. Do not use the '!=' operator, as it does not accurately represent the exclusion of the specific condition.

Question: Find the last name of students who is either female (sex is F) and living in the city of code BAL or male (sex is M) and in age of below 20.,
Correct SQL: SELECT lname FROM student WHERE sex  =  'F' AND city_code  =  'BAL' UNION SELECT lname FROM student WHERE sex  =  'M' AND age  <  20,
Erroneous SQL: SELECT LName FROM Student WHERE (Sex = 'F' AND city_code = 'BAL') OR (Sex = 'M' AND Age < 20)
Please write an instruction to tell GPT4 do not make this mistake again.
Instruction: When translating a question that requires finding elements that meet either of two distinct conditions, use the UNION operator in SQL to combine the results of two separate SELECT statements. Do not use the OR operator within a single WHERE clause, as it does not accurately represent the distinct conditions.

Question: How many exhibitions has each artist had?,
Correct SQL: SELECT T2.name ,  count(*) FROM exhibition AS T1 JOIN artist AS T2 ON T1.artist_id  =  T2.artist_id GROUP BY T1.artist_id,
Erroneous SQL: SELECT artist.Name, COUNT(exhibition.Exhibition_ID) AS Number_of_Exhibitions FROM artist LEFT JOIN exhibition ON artist.Artist_ID = exhibition.Artist_ID GROUP BY artist.Name
Please write an instruction to tell GPT4 do not make this mistake again.
Instruction: When translating a question that requires counting the number of occurrences for each category, make sure to group by the ID, not the name, use join instead of left join.

Question: Find the three most expensive procedures.,
Correct SQL: SELECT name FROM procedures ORDER BY cost LIMIT 3,
Erroneous SQL: SELECT Name, Cost FROM Procedures ORDER BY Cost DESC LIMIT 3
Please write an instruction to tell GPT4 do not make this mistake again.
Instruction: When translating a question that requires finding the top N elements based on a certain attribute, ensure that the SQL query only selects the required attribute, not additional ones. In this case, only the 'name' attribute should be selected, not 'Cost'. Also, remember to order the results in descending order to get the most expensive procedures.

Question: List the name, born state and age of the heads of departments ordered by age.,
Correct SQL: SELECT name ,  born_state ,  age FROM head ORDER BY age,
Erroneous SQL: SELECT head.name, head.born_state, head.age FROM head INNER JOIN management ON head.head_ID = management.head_ID ORDER BY head.age
Please write an instruction to tell GPT4 do not make this mistake again.
Instruction: 
\end{lstlisting}

\subsection{EPI-context Prompt}
The EPI-context prompt generates an EPI for the current task, and the demonstrations in the EPI-context prompt are task-dependent.
\begin{lstlisting}[xleftmargin=0pt]
Question: "Find the names and number of works of the three artists who have produced the most songs.",
Instruction: When translating a question that requires finding the top N elements based on a count of a certain attribute, ensure that the SQL query uses the COUNT(*) function and orders by this count in descending order. Also, remember to group by the correct attribute. In this case, the query should group by 'artist_name' from the 'song' table, not the 'artist' table.

Question: "How many exhibitions has each artist had?",
Instruction: When translating a question that requires counting the number of occurrences for each category, make sure to group by the ID, not the name. Also, use JOIN instead of LEFT JOIN to ensure that only artists who have had exhibitions are included in the count.

Question: "How many customers are there?",
Instruction: When translating a question that requires counting the total number of a certain entity, ensure that the SQL query is correctly targeting the right table and column. In this case, the total number of customers is stored in the 'no_of_customers' column in the 'bank' table, not in a 'customer' table. Always make sure to understand the database schema correctly before translating the question.

Question: "How many players are from each country?",
Instruction: When translating a question that requires counting the number of occurrences for each category, ensure that the SQL query only includes the necessary tables in the JOIN clause. In this case, the 'player' table is not needed to answer the question. The query should only JOIN the 'country' and 'match_season' tables.

Question: "Please show the nominee who has been nominated the greatest number of times.",
Instruction: When translating a question that requires finding the element with the highest count, ensure that the SQL query only selects the required attribute, not additional ones. In this case, only the 'Nominee' attribute should be selected, not 'Nomination_Count'. Also, remember to order the results in descending order to get the nominee with the greatest number of nominations.

Question: "Show all artist names and the number of exhibitions for each artist.",
Instruction: When translating a question that requires counting the number of occurrences for each category, make sure to group by the ID, not the name. Also, use JOIN instead of LEFT JOIN to ensure that only artists who have had exhibitions are included in the results.

Question: "What is the count of the songs that last approximately 4 minutes?",
Instruction: When translating a question that requires finding elements based on an approximate condition, use the LIKE operator in SQL. Do not use the '=' operator, as it does not accurately represent the approximation. In this case, the duration should be represented as "4

Question: "What is the most popular first name of the actors?",
Instruction: When translating a question that requires finding the most frequent occurrence of a certain attribute, ensure that the SQL query only selects the required attribute, not additional ones. In this case, only the 'first_name' attribute should be selected, not 'count'. Also, remember to order the results in descending order to get the most popular first name.

Question: "How many kinds of roles are there for the staff?",
Instruction: When translating a question that requires counting distinct elements, ensure that the SQL query is selecting from the correct table. In this case, the table should be 'Project_Staff', not 'Staff_Roles'. Always double-check the table names in the question and make sure they match with the ones in your SQL query.

Question: "Show all party names and the number of members in each party.",
Instruction: When translating a question that requires counting the number of occurrences for each category, make sure to group by the ID, not the name, use join instead of left join. Also, ensure that the SQL query only selects the required attributes, not additional ones.

Question: "What is the total number of singers?",
Instruction:
\end{lstlisting}

\subsection{SQLGen+EPI Prompt}
The SQLGen+EPI prompt is used to generate SQL with the assistance of EPI.

\begin{lstlisting}[xleftmargin=0pt]
CREATE TABLE stadium ( 
Stadium_ID int,
Location text,
Name text,
Capacity int, 
Highest int,
Lowest int, 
Average int, 
PRIMARY KEY (Stadium_ID)
);

CREATE TABLE singer ( 
Singer_ID int,
Name text,
Country text, 
Song_Name text, 
Song_release_year text, 
Age int, 
Is_male bool, 
PRIMARY KEY (Singer_ID)
);

CREATE TABLE concert ( 
concert_ID int, 
concert_Name text, 
Theme text, 
Stadium_ID int, 
Year int,
PRIMARY KEY (concert_ID),
FOREIGN KEY (Stadium_ID) REFERENCES stadium(Stadium_ID)
);

CREATE TABLE singer_in_concert ( 
concert_ID int, 
Singer_ID int, 
PRIMARY KEY (concert_ID,Singer_ID),
FOREIGN KEY (concert_ID) REFERENCES concert(concert_ID),
FOREIGN KEY (Singer_ID) REFERENCES singer(Singer_ID)
);
Question: "What is the total number of singers?",
Please translate this Question to a SQL.
When translating a question that requires counting the total number of a certain entity, ensure that the SQL query is correctly targeting the right table and column. In this case, the total number of singers is stored in the 'singer' table, not in a 'song' table. Always make sure to understand the database schema correctly before translating the question.
SQL:
\end{lstlisting}

\section{Examples in QSESet}
These are some of the examples in the QSESet, where the EPIs are generated by the EPIGen prompt, i.e. by comparing the erroneous SQL with the gold SQL.

\begin{lstlisting}[xleftmargin=0pt]
Question: List the most common type of Status across cities.",

Gold SQL: SELECT Status FROM city GROUP BY Status ORDER BY COUNT(*) DESC LIMIT 1,

Erroneous SQL: SELECT Status, COUNT(Status) AS Frequency FROM city GROUP BY Status ORDER BY Frequency DESC LIMIT 1,
 
EPI: When translating a question that requires finding the most common element based on a certain attribute, ensure that the SQL query only selects the required attribute, not additional ones. In this case, only the 'Status' attribute should be selected, not 'Frequency'. Also, remember to use COUNT(*) in the ORDER BY clause to correctly count the occurrences of each status.,

EPI-SQL: SELECT Status FROM city GROUP BY Status ORDER BY COUNT(*) DESC LIMIT 1
\end{lstlisting}

\begin{lstlisting}[xleftmargin=0pt]
Question: From which hometowns did no gymnasts come from?,

Gold SQL: SELECT DISTINCT Hometown FROM people EXCEPT SELECT DISTINCT T2.Hometown FROM gymnast AS T1 JOIN people AS T2 ON T1.Gymnast_ID  =  T2.People_ID,

Erroneous SQL: SELECT Hometown FROM people WHERE People_ID NOT IN (SELECT Gymnast_ID FROM gymnast),

EPI: When translating a question that requires finding elements that are not present in a certain condition, use the EXCEPT operator in SQL. Do not use the NOT IN operator, as it does not accurately represent the exclusion of the specific condition. Also, remember to use DISTINCT to avoid duplicate hometowns.,

EPI-SQL: SELECT DISTINCT Hometown FROM people EXCEPT SELECT DISTINCT Hometown FROM people WHERE People_ID IN (SELECT Gymnast_ID FROM gymnast)
\end{lstlisting}

\begin{lstlisting}[xleftmargin=0pt]
Question: What are the names of reviewers who had rated 3 star and 4 star?,

Gold  SQL: SELECT T2.name FROM Rating AS T1 JOIN Reviewer AS T2 ON T1.rID  =  T2.rID WHERE T1.stars  =  3 INTERSECT SELECT T2.name FROM Rating AS T1 JOIN Reviewer AS T2 ON T1.rID  =  T2.rID WHERE T1.stars  =  4,

Erroneous SQL: SELECT Reviewer.name FROM Reviewer JOIN Rating ON Reviewer.rID = Rating.rID WHERE Rating.stars IN (3, 4) GROUP BY Reviewer.name,

EPI: When translating a question that requires finding common elements based on two distinct conditions, use the INTERSECT operator in SQL. Do not use the IN operator within a WHERE clause and GROUP BY, as it does not accurately represent the intersection of the two conditions.,

EPI-SQL: SELECT name FROM Reviewer WHERE rID IN (SELECT rID FROM Rating WHERE stars = 3) INTERSECT SELECT name FROM Reviewer WHERE rID IN (SELECT rID FROM Rating WHERE stars = 4)
\end{lstlisting}

\begin{lstlisting}[xleftmargin=0pt]
Question: Find the names of courses that have either 3 credits or 1 credit but 4 hours.,

Gold SQL: SELECT CName FROM COURSE WHERE Credits  =  3 UNION SELECT CName FROM COURSE WHERE Credits  =  1 AND Hours  =  4,

Erroneous SQL: SELECT CName FROM Course WHERE (Credits = 3 OR Credits = 1) AND Hours = '4',

EPI: When translating a question that requires finding elements that meet either of two distinct conditions, use the UNION operator in SQL to combine the results of two separate SELECT statements. Do not use the OR operator within a single WHERE clause, as it does not accurately represent the distinct conditions. In this case, the condition for '1 credit' should only be applied when 'Hours' equals '4', not for all courses.,

EPI-SQL: SELECT CName FROM Course WHERE Credits = 3 UNION SELECT CName FROM Course WHERE Credits = 1 AND Hours = '4'
\end{lstlisting}

\section{Examples of EPI and EPI-SQL}
We generate an EPI that matches the current task and use the EPI to guide the LLMs in solving the current task, some examples of which are shown below. The EPIs here are generated using the EPI-context prompt, which is generated without the involvement of erroneous SQL.

\begin{lstlisting}[xleftmargin=0pt]
Question: Find the first name of students who have cat or dog pet.

Gold SQL: SELECT DISTINCT T1.Fname FROM student AS T1 JOIN has_pet AS T2 ON T1.stuid  =  T2.stuid JOIN pets AS T3 ON T3.petid  =  T2.petid WHERE T3.pettype  =  'cat' OR T3.pettype  =  'dog'

ZeroShot SQL: SELECT Student.Fname FROM Student JOIN Has_Pet ON Student.StuID = Has_Pet.StuID JOIN Pets ON Has_Pet.PetID = Pets.PetID WHERE Pets.PetType IN ('cat', 'dog')

EPI: When translating a question that requires finding elements that meet any of multiple conditions, use the OR operator in SQL to find the elements that satisfy any of the conditions. Do not use the UNION operator, as it does not accurately represent the union of the conditions.

EPI-SQL: SELECT Student.Fname FROM Student JOIN Has_Pet ON Student.StuID = Has_Pet.StuID JOIN Pets ON Has_Pet.PetID = Pets.PetID WHERE Pets.PetType = 'cat' OR Pets.PetType = 'dog'
\end{lstlisting}

\begin{lstlisting}[xleftmargin=0pt]
Question: What are the id and names of the countries which have more than 3 car makers or produce the 'fiat' model?

Gold SQL: SELECT T1.countryId ,  T1.CountryName FROM Countries AS T1 JOIN CAR_MAKERS AS T2 ON T1.CountryId  =  T2.Country GROUP BY T1.countryId HAVING count(*)  >  3 UNION SELECT T1.countryId ,  T1.CountryName FROM Countries AS T1 JOIN CAR_MAKERS AS T2 ON T1.CountryId  =  T2.Country JOIN MODEL_LIST AS T3 ON T2.Id  =  T3.Maker WHERE T3.Model  =  'fiat';

ZeroShot SQL: SELECT c.CountryId, c.CountryName FROM countries c JOIN car_makers cm ON c.CountryId = cm.Country JOIN model_list ml ON cm.Id = ml.Maker WHERE cm.Maker IN (SELECT cm.Maker FROM car_makers cm GROUP BY cm.Maker HAVING COUNT(cm.Id) > 3) OR ml.Model = 'fiat'

EPI: When translating a question that requires finding elements based on a count condition or a specific attribute, use the UNION operator in SQL to combine the results from two separate SELECT statements. Do not use the OR operator in the WHERE clause, as it does not accurately represent the union of the two conditions. Also, remember to group by the ID, not the name, when counting the number of car makers.

EPI-SQL: SELECT CountryId, CountryName FROM countries WHERE CountryId IN ( SELECT Country FROM car_makers GROUP BY Country HAVING COUNT(Id) > 3 ) UNION SELECT CountryId, CountryName FROM countries WHERE CountryId IN ( SELECT Country FROM car_makers WHERE Id IN ( SELECT Maker FROM model_list WHERE Model = 'fiat' ) )
\end{lstlisting}

\begin{lstlisting}[xleftmargin=0pt]
Question: Which airline has most number of flights?

Gold SQL: SELECT T1.Airline FROM AIRLINES AS T1 JOIN FLIGHTS AS T2 ON T1.uid  =  T2.Airline GROUP BY T1.Airline ORDER BY count(*) DESC LIMIT 1

ZeroShot SQL: SELECT a.Airline, COUNT(*) as NumberOfFlights FROM airlines a JOIN flights f ON a.uid = f.Airline GROUP BY a.Airline ORDER BY NumberOfFlights DESC LIMIT 1

EPI: When translating a question that requires finding the element with the maximum count, ensure that the SQL query only selects the required attribute, not additional ones. In this case, only the 'airline' attribute should be selected, not 'number_of_flights'. Also, remember to order the results in descending order to get the airline with the most number of flights.

EPI-SQL: SELECT a.Airline FROM airlines a JOIN flights f ON a.uid = f.Airline GROUP BY a.Airline ORDER BY COUNT(*) DESC LIMIT 1
\end{lstlisting}

\begin{lstlisting}[xleftmargin=0pt]
Question: Find the first name, country code and birth date of the winner who has the highest rank points in all matches.

Gold SQL: SELECT T1.first_name ,  T1.country_code ,  T1.birth_date FROM players AS T1 JOIN matches AS T2 ON T1.player_id  =  T2.winner_id ORDER BY T2.winner_rank_points DESC LIMIT 1

ZeroShot SQL: SELECT players.first_name, players.country_code, players.birth_date FROM players JOIN matches ON players.player_id = matches.winner_id WHERE matches.winner_rank_points = (SELECT MAX(winner_rank_points) FROM matches)

EPI: When translating a question that requires finding the maximum value of an attribute and returning corresponding values, use the ORDER BY and LIMIT 1 clauses in SQL. Do not use subqueries with the MAX function, as they do not accurately represent the requirement of finding the winner with the highest rank points. Also, ensure to join the relevant tables correctly to get the accurate result.

EPI-SQL: SELECT players.first_name, players.country_code, players.birth_date FROM players JOIN matches ON players.player_id = matches.winner_id ORDER BY matches.winner_rank_points DESC LIMIT 1
\end{lstlisting}

\section{Error Distribution and Error Rate for Databases}
\begin{figure*}[!ht]
    \centering
    \includegraphics[width=\linewidth,scale=1]{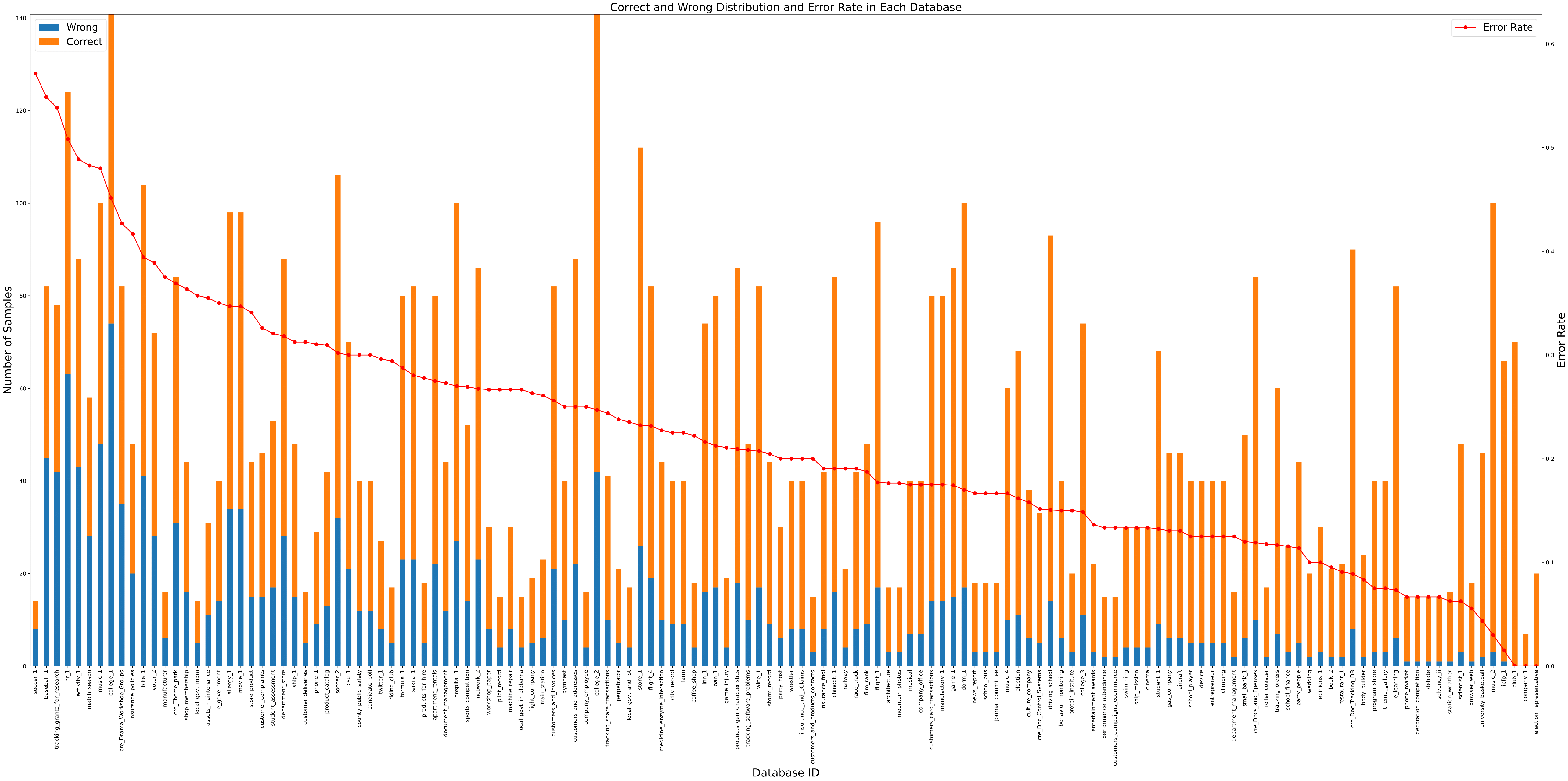}
    \caption{Error distribution and error rate for all databases.}
    \label{fig:bias of all schema}
\end{figure*}

In Figure \ref{fig:bias of schema}, we filtered out the databases with less than 50 samples, and the error distribution and error rate of the samples of all the databases are shown in Figure \ref{fig:bias of all schema}.

\end{document}